\documentclass[runningheads]{llncs}
\usepackage[T1]{fontenc}
\usepackage{graphicx}
\usepackage{xcolor}
\usepackage{amsmath}
\usepackage{enumitem}
\usepackage{amsfonts}
\usepackage{subcaption}
\usepackage{tikz}
\usepackage{multirow}
\usepackage{ragged2e}
\usetikzlibrary{positioning, arrows.meta, shapes.geometric}

\begin{document}

\title{Multimodal Knowledge Distillation for Gastric Adenocarcinoma Classification from Whole-Slide Images}
\titlerunning{Multimodal Knowledge Distillation for Gastric Cancer Classification}

\author{Shrihari Dumbre\inst{1} \and
Bikash Santra\inst{1}}

\institute{Indian Institute of Technology, Jodhpur \\
\email{b23ee1069@iitj.ac.in}, \email{bikash@iitj.ac.in}}
\maketitle              
\begin{abstract}

Gastric adenocarcinoma (GA) is a leading cause of cancer-related mortality worldwide, and accurate histopathological subtype classification from whole-slide images (WSIs) is essential for effective treatment planning. While multimodal approaches that integrate pathology report text with WSIs can improve classification, existing methods often depend on computationally expensive transformer architectures and large language models. We propose a multimodal knowledge distillation (MKD) framework that combines a pretrained WSI image encoder and a clinical text encoder using Low-Rank Multimodal Fusion (LMF) to efficiently model cross-modal interactions during training. Each WSI is represented as a bag of patches paired with a slide-level diagnostic caption. The teacher model learns fused image–text representations for subtype classification, while the student model distills this knowledge to enable accurate image-only inference. We evaluate our method on the PatchGastric benchmark dataset and achieve at least 3.35\% higher mean accuracy than state-of-the-art approaches, without relying on transformer-based fusion, multi-task learning, or large language models. The source code is available at \url{https://github.com/helomelo1/MKD-LMF}.

\keywords{Multimodal Learning \and Low-Rank Multimodal Fusion \and Knowledge Distillation \and Gastric Cancer Classification \and Histopathology Image Analysis}
\end{abstract}

\section{Introduction}

Accurate histological subtype classification of gastric adenocarcinoma from whole-slide images (WSIs) is clinically critical for treatment planning, yet morphological similarity between subtypes makes this task challenging even for experienced pathologists~\cite{patchgastric}. Most recent multimodal approaches~\cite{pathomicfusion,mcat,motcat,cmta,pathomics,pathm3} demonstrate that pairing WSI features with complementary data sources, such as genomic sequencing data, gene expression profiles and clinical records, improves classification. But most of them either require genomic sequencing data that is unavailable in routine clinical workflows~\cite{pathomicfusion,mcat} or rely on highly complex architectures, such as PathM3~\cite{pathm3}. It employs a 12-layer Query Transformer initialized from BLIP-2~\cite{blip2} together with a frozen FlanT5-XL decoder~\cite{chung2022flan} for classification and caption generation jointly. 

The PatchGastric dataset~\cite{patchgastric} is one of the few publicly available datasets that provide both histopathology patch images and WSI-level diagnostic captions (i.e., report summaries), making it well-suited for multimodal analysis of gastric adenocarcinoma. However, the dataset presents a significant challenge. Many WSIs across different subtypes share near-identical descriptions, limiting their standalone discriminating power. Due to real-world data scarcity, methods on this dataset follow a data-scarce training protocol, using only 20\% of the WSIs for training. Prior methods on this dataset, such as PathM3~\cite{pathm3} address these challenges using complex multi-task learning-based frameworks that jointly optimize classification and caption-generation using large-language models. While effective, these methods introduce substantial computational overhead.

On the contrary, we propose an LMF-based multimodal fusion approach that fuses patch-level image features with pathology report captions using Low-Rank Multimodal Fusion (LMF)~\cite{lmf}. 
Notably, each WSI in the PatchGastric dataset~\cite{patchgastric} is provided as a set of pre-extracted patches of size $300 \times 300$ at $20\times$ magnification. Following the Multiple Instance Learning paradigm~\cite{abmil}, each WSI is treated as a bag of patch instances. Each patch embedding is obtained and mean-pooled via LMF to fuse with the WSI-level caption embedding. LMF models multiplicative cross-modal interactions through a low-rank tensor decomposition without requiring attention-based cross-modal modules~\cite{mcat,pathm3} or generative objectives~\cite{blip2,pathm3}. Pathology report captions are routinely generated during standard diagnostic workflows, making them a practical, readily available second modality alongside H\&E-stained histopathology images. Unlike PG-MLIF~\cite{pgmlif}, which applies LMF to fuse WSIs with genomic data for survival prediction, we apply LMF to the image-report setting for histological subtype classification of gastric adenocarcinoma.


Our experiments compare the proposed method against 
ABMIL~\cite{abmil}, CLAM~\cite{clam}, 
DSMIL~\cite{dsmil}, TransMIL~\cite{transmil}, 
CITE~\cite{cite}, ILRA-MIL~\cite{ilramil} 
and PathM3~\cite{pathm3}. 
We use the same PatchGastric dataset and the same train-validation-test split ratio as in CITE~\cite{cite} and PathM3~\cite{pathm3}. We use three gastric adenocarcinoma subtypes, including “well differentiated tubular adenocarcinoma”, “moderately differentiated tubular adenocarcinoma”, and “poorly differentiated adenocarcinoma”, similar to these baselines, thus enabling a direct comparison. We further conduct experiments on different multimodal fusion strategies and ablation studies across multiple LMF rank settings. 
The \textbf{contributions} of this work are the following:
\begin{itemize}[topsep=0pt]


    \item \textbf{Lightweight multimodal framework:} We propose an efficient multimodal knowledge distillation (MKD) approach that fuses WSI features and clinical text using Low-Rank Multimodal Fusion (LMF), avoiding heavy transformer-based architectures and large language models.
    
    \item \textbf{Knowledge distillation for image-only inference:} We introduce a teacher –student setup where multimodal knowledge learned during training is distilled into a student model, enabling accurate and efficient inference using only WSIs.

    \item \textbf{Improved performance with efficiency:} We conduct extensive experiments to analyze the contributions of individual modalities and the effect of the LMF rank in improving the gastric adenocarcinoma classification performance.
\end{itemize}

\section{Related Works}
\subsection{Multiple Instance Learning for Histopathology}
Multiple Instance Learning (MIL) is a weakly supervised paradigm where training examples are organized into bags of instances, with labels available only at the bag level~\cite{dietterich1997}. In computational pathology, a whole-slide image (WSI) is treated as a bag containing hundreds to thousands of patch-level instances (small image regions extracted at specific magnification), with the number of patches varying across WSIs. The WSI-level label is known, but individual patch labels are not, making MIL a natural fit for WSI analysis. Ilse et al.~\cite{abmil} proposed Attention-based MIL (ABMIL), which learns attention weights to aggregate patch features into a bag-level representation. In another work, Shao et al.~\cite{transmil} introduced Transformer-based correlated MIL (TransMIL), employing transformer-based self-attention with Nystr\"om approximation~\cite{xiong2021} to model correlations among patches efficiently. Whereas, Lu et al.~\cite{clam} proposed Clustering-Constrained-Attention Multiple-Instance Learning (CLAM), a data-efficient and weakly supervised framework for WSI classification. While Li et al.~\cite{dsmil} introduced Dual-Stream MIL (DSMIL), a dual-stream network that incorporates self-supervised contrastive learning. All of these methods rely exclusively on image features for histopathology classification. On the contrary, we introduce an MIL-based multimodal framework that uses a bag-of-image-patches representation and a caption generated from the pathology report.

\subsection{Multimodal Fusion on Computation Pathology} A number of recent works~\cite{pathomicfusion,mcat,motcat,cmta,pathomics,pathm3} have explored multimodal fusion for cancer prognosis and classification by combining histopathology images along with complementary data modalities. Chen et al.~\cite{pathomicfusion} proposed Pathomic Fusion, integrating histopathology and genomic features via the Kronecker product for survival prediction. While effective in modelling cross-modal relationships, its tensor fusion is computationally expensive, requires paired genomic data, and assumes the availability of both modalities during inference. 
In another work, Chen et al.~\cite{mcat} introduced Multimodal Co-Attention Transformer (MCAT), a co-attention Transformer fusing WSIs with genomic data, while Xu and Chen~\cite{motcat} proposed Multimodal Optimal Transport Co-Attention Transformer (MOTCat) using optimal transport-based co-attention. On the other hand, Zhou and Chen~\cite{cmta} introduced Cross-Modal Translation and Alignment (CMTA) for cross-modal translation and alignment, and Ding et al.~\cite{pathomics} proposed PathOmics, a pathology-genomics Transformer for survival prediction. On the other hand, recent vision-language approaches focus on image-text representations that leverage widely available diagnostic captions, enabling more flexible multimodal learning. For example, most closely related to our work, Zhou et al.~\cite{pathm3} proposed PathM3, an MIL-based multimodal framework. PathM3 aligns WSIs with diagnostic captions using a query-based transformer initialized with BLIP-2~\cite{blip2}, jointly training for classification and caption generation. While PathM3 achieves strong performance, it relies on a computationally expensive architecture with a 12-layer query Transformer and a frozen FlanT5-XL decoder. In contrast, we employ a knowledge distillation framework with a low-rank multimodal fusion approach that directly learns joint image-text representations for WSI-only classification, avoiding the need for caption-generation objectives and large language models while achieving superior classification performance. 

\subsection{Low-Rank Multimodal Fusion and Knowledge Distillation} Low-Rank Multimodal Fusion (LMF) was introduced by Liu et al.~\cite{lmf} as an efficient alternative to Tensor Fusion Networks~\cite{tfn}. LMF uses parallel low-rank decomposition to model pairwise interactions between features from the image and text modalities while avoiding the exponential parameter growth of full tensor fusion~\cite{tfn}. For cancer survival prediction, Pan et al.~\cite{pgmlif} applied LMF to fuse histopathology images with genomic data. PG-MLIF~\cite{pgmlif} requires genomic profiling data, which is expensive and not always clinically available. Unlike PG-MLIF~\cite{pgmlif}, we apply LMF to fuse histopathology images with pathology report summaries, which are routinely generated as part of standard diagnostic workflows.
Knowledge distillation~\cite{hinton2015} transfers knowledge from a large teacher model to a smaller student model. FitNets~\cite{romero2015} extended this by distilling intermediate representations, thereby enabling richer feature learning. In our approach, we employ knowledge distillation to transfer the multimodal representation learned from both image and text to a student model that operates solely on images.

\section{Methodology}

\begin{figure}[t]
    \centering
    \includegraphics[
        width=\linewidth,
    ]{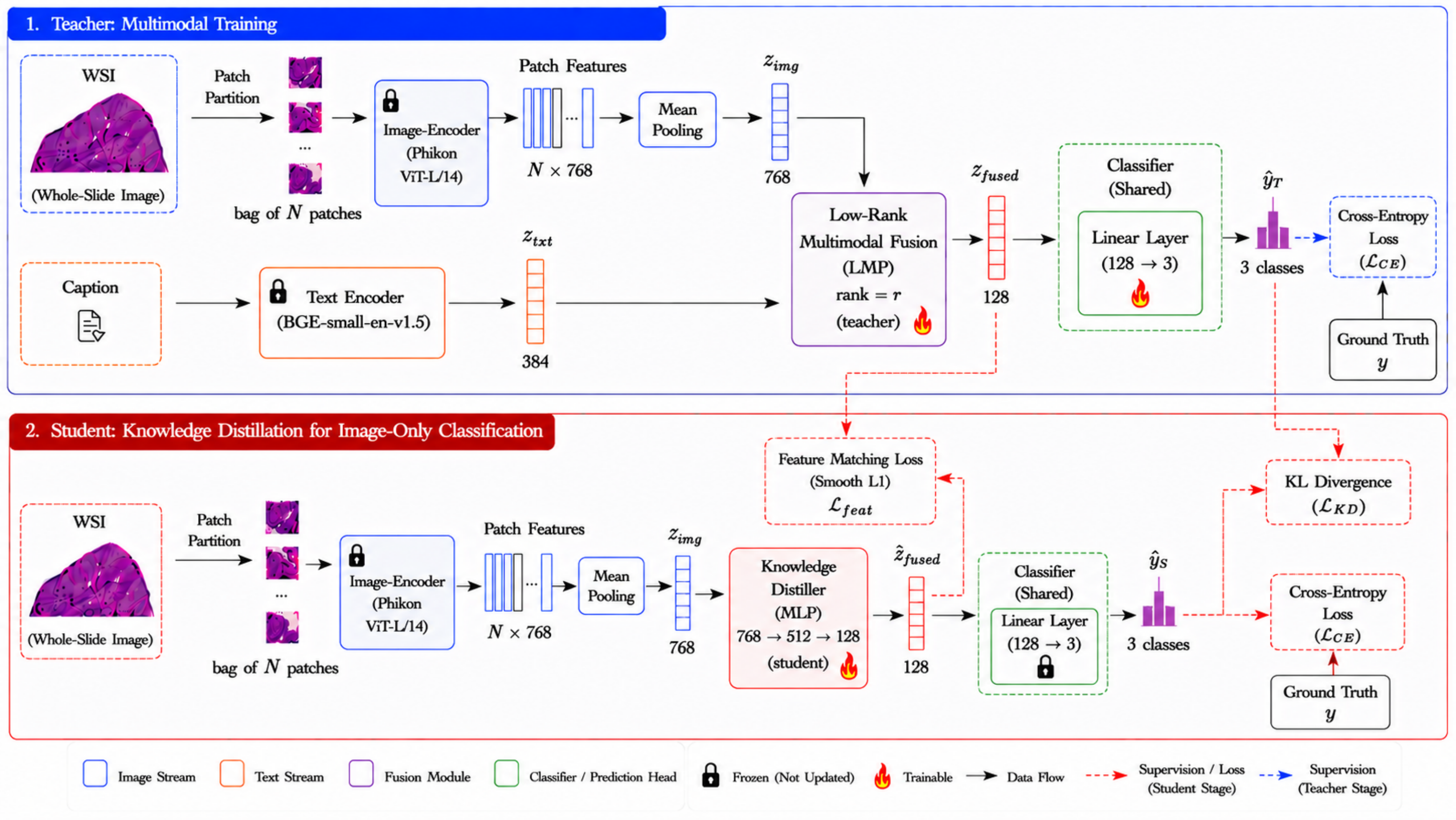}
    \caption{
    {Overview of the proposed two-stage MKD-LMF framework.} WSI patches are encoded and mean-pooled to obtain a bag-level image representation $(\mathbf{z}_{img})$, while the corresponding caption is passed through a text encoder to extract text features $(\mathbf{z}_{txt})$. In \textbf{Stage 1}, the \textbf{teacher network} employs a Low-Rank Multimodal Fusion (LMF) module to combine $(\mathbf{z}_{img})$ and $(\mathbf{z}_{txt})$, producing a fused representation $(\mathbf{z}_{fused})$, which is used to train a linear classifier (shared with the student) for gastric adenocarcinoma (GA) subtype classification. In \textbf{Stage 2}, the \textbf{student network} takes only $(\mathbf{z}_{img})$ as input and learns, via a linear knowledge distillation module, to approximate $\mathbf{z}_{fused}$ generated by the teacher’s frozen LMF module. The synthesized representation is then passed to the shared classifier to enable \textbf{image-only GA subtype classification}.
    }
    \label{fig:architecture}
\end{figure}

\subsection{Problem Formulation}
Consider a dataset of $M$ whole-slide images (WSIs), where each WSI is represented as a bag of $N$ patches $\mathcal{B} = \{ {p_{1}, p_{2}, \ldots, p_{N}} \}$, with $N$ varying across WSIs based on tissue regions. Each WSI is paired with a diagnostic caption $c$ provided in the dataset, derived from the corresponding pathology report. The goal is to classify WSIs into $K = 3$ subtypes of gastric adenocarcinoma: well-differentiated, moderately differentiated, and poorly differentiated tubular adenocarcinoma.

We address this 3-way classification of gastric adenocarcinoma in two stages. First, a multimodal teacher $f_{\text{teacher}}: (\mathcal{B}, c) \rightarrow y$ is trained using both image and text modalities. Second, an image-only student $f_{\text{student}}: \mathcal{B} \rightarrow y$ is trained via knowledge distillation, enabling WSI-only inference without the pathology report. The proposed complete pipeline is illustrated in Fig. \ref{fig:architecture}.

\subsection{Image Feature Extraction} \label{image_feature}
First, the patches of a WSI $p_{i} \in \mathcal{B} \subset \mathbb{R}^{H \times W \times 3}$ with height $H$ and width $W$ are resized to $224 \times 224$. Subsequently, each patch $p_{i}$ is passed through a pretrained image encoder $\phi_{\text{img}}$ to obtain a $d_v$-dimensional patch-level image embedding:
\begin{equation}
    \mathbf{v}_{i} = \phi_{\text{img}}(p_{i}) \in \mathbb{R}^{d_v}.
\end{equation}
In our implementation, Phikon~\cite{phikon}, a Vision Transformer (ViT-B/16)~\cite{vit} pretrained on large-scale histopathology data via self-supervised learning~\cite{mae} is used as our image encoder, where $d_v = 768$. The weights of $\phi_{\text{img}}$ are kept frozen during training of the proposed model.

\subsection{Bag-Level Feature Representation via Aggregation} \label{aggregator}
Following the MIL paradigm, we aggregate the patch-level representations $\mathbf{v}_{i}$ within each bag $\mathcal{B}$ to obtain a single bag-level representation. Since patches within a bag have no inherent ordering, the aggregation function must be permutation invariant, meaning the output should not change when the order of input patches is rearranged. This property is essential because the model should produce the same prediction regardless of how patches are arranged. Thus, we use mean pooling as the permutation-invariant aggregation function:
\begin{equation}
    \mathbf{z}_{img} = \frac{1}{N} \sum_{k=1}^{N} \mathbf{v}_{k} \in \mathbb{R}^{d_v},
\end{equation}
to treat all patches equally, providing a simple yet effective summarization of the bag contents.

\subsection{Text Feature Extraction}
The caption $c$, a summarized pathology report, associated with the input WSI is encoded using a pretrained text encoder $\phi_{\text{txt}}$ to obtain a $d_t$-dimensional bag-level text embedding:
\begin{equation}
    \mathbf{z}_{txt} = \phi_{\text{txt}}(c) \in \mathbb{R}^{d_t}.
\end{equation}
We use BGE-small-en-v1.5~\cite{bge}, a pretrained sentence-embedding model, as our text encoder, with $d_t = 384$. Since all patches within a bag share the same WSI-level caption, the text embedding $\mathbf{z}_{txt}$ is paired with the bag-level aggregated and fixed-dimensional representation $\mathbf{z}_{img}$ of WSI patches. The weights of $\phi_{\text{txt}}$ are kept frozen during the training of our model.

\subsection{Low-Rank Multimodal Fusion}
Multimodal GA subtype classification requires capturing high-order correlations between visual and textual embeddings $\mathbf{z}_{img}$ and $\mathbf{z}_{txt}$. Simple additive fusion schemes (such as feature concatenation or summation) assume that modalities contribute independently, which fails to model inter-modal dependencies. Conversely, multiplicative fusion captures the cross-modal interactions by computing pairwise products between every image feature and every 
text feature. However, mapping this full tensor to the output space requires a high-dimensional fusion weight tensor whose parameters scale quadratically with the number of features, leading to severe memory overhead. To capture and approximate multiplicative cross-modal interactions between the image and text modalities, we employ Low-Rank Multimodal Fusion (LMF)~\cite{lmf}. 

Given a bag-level representation of patches $\mathbf{z}_{img} \in \mathbb{R}^{d_v}$ and the caption embedding $\mathbf{z}_{txt} \in \mathbb{R}^{d_t}$, LMF computes a fused representation. It introduces two sets of learnable projection matrices $\mathbf{W}_v^{(k)}$ and $\mathbf{W}_t^{(k)}$ specific to the image and text modalities, respectively
We append a constant 1 to each embedding vector to incorporate a bias term into the low-rank factorization, enabling the model 
to learn an additive offset without a separate bias parameter. Appending a constant 1 allows the augmented dimensions to act as multiplicative identities, enabling the factorization to represent image-only, text-only, and bias terms in addition to cross-modal interactions as follows:
\begin{equation}
    \bar{\mathbf{z}}_{img} = [\mathbf{z}_{img}; 1] \in \mathbb{R}^{d_v + 1}, \quad \bar{\mathbf{z}}_{txt} = [\mathbf{z}_{txt}; 1] \in \mathbb{R}^{d_t + 1}.
\end{equation}

Rather than explicitly constructing the full fusion tensor, LMF factorizes the fusion weight tensor into $r$ distinct components. For each factor $j \in \{1, \ldots, r\}$, learnable projection matrices $\mathbf{W}_v^{(j)} \in \mathbb{R}^{d_f \times (d_v+1)}$ and $\mathbf{W}_t^{(j)} \in \mathbb{R}^{d_f \times (d_t+1)}$ project the augmented embeddings into a common $d_f$-dimensional space, and the fused representation is obtained by summing element-wise products across all factors:
\begin{equation}
    \mathbf{z}_{fused} = \sum_{j=1}^{r} \left( \mathbf{W}_v^{(j)} \bar{\mathbf{z}}_{img} \right) \odot \left( \mathbf{W}_t^{(j)} \bar{\mathbf{z}}_{txt} \right) \in \mathbb{R}^{d_f},
\end{equation}
where $\odot$ denotes the Hadamard (element-wise) product, $r$ is the rank, a hyperparameter controlling the expressiveness of the decomposition, and $d_f$ is the fused embedding dimension. This formulation effectively retains the rich cross-modal interactions. Conventional multiplicative fusion requires $\mathcal{O}(d_f \cdot d_v \cdot d_t)$ trainable parameters. In contrast, LMF reduces this parameter count to $\mathcal{O}(r \cdot d_f \cdot (d_v + d_t))$, and thereby drastically improves parameter efficiency. 

\subsection{Classification} \label{classifier}
The proposed teacher-student MKD framework uses a shared classifier trained on the teacher side, which is kept frozen on the student side to distil the teacher's knowledge. As shown in Fig. \ref{fig:architecture}, on the teacher network, the fused multimodal representation $\mathbf{z}_{fused}$ is fed to a linear classifier ($\phi_{\text{cls}}$) to obtain the predicted GA subtype class probabilities:
\begin{equation}
    \hat{\mathbf{y}}_T = \phi_{\text{cls}}(\mathbf{z}_{fused})\in \mathbb{R}^{K}.
\end{equation}
\begin{figure}[t]
    \centering
    \includegraphics[
        width=11cm,
    ]{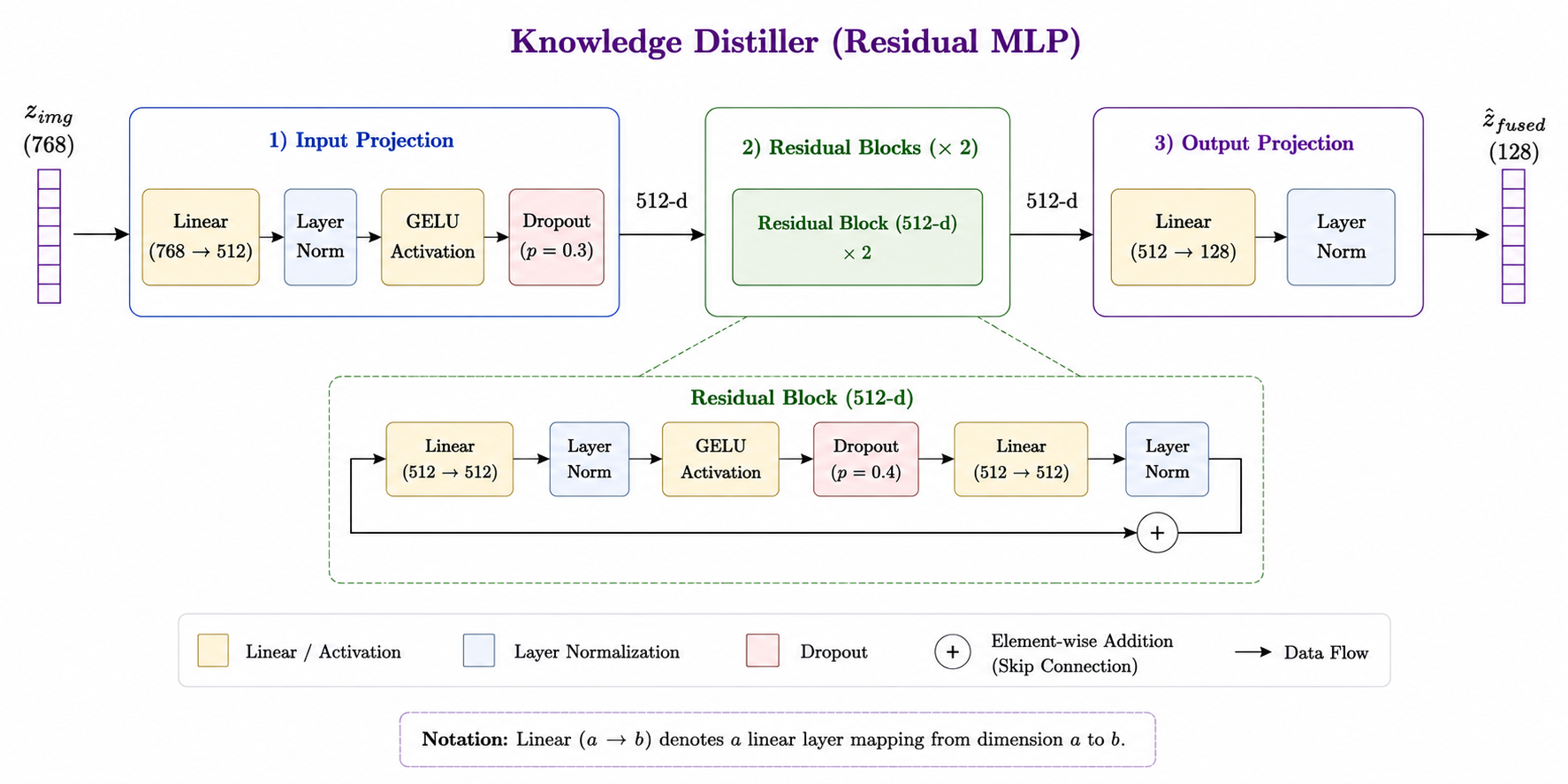}
    \caption{Architecture of the proposed Knowledge Distiller, a residual multi-layer perceptron (MLP), comprising an input projection, stacked residual blocks, and an output projection for mapping image features to the fused feature space.
    }
    \label{fig:kd_architecture}
\end{figure}
\subsection{Knowledge Distillation for Synthesizing Multimodal Embedding} \label{KD}
The multimodal model often achieves high accuracy by using both image and text data. However, it requires the pathology report at inference time. Pathology reports for histopathology images are generated after pathologists have made the diagnostic decision and identified the GA subtype. This limits its clinical applicability, as our objective is to automate this GA subtype identification. To address this limitation, we introduce a student network in which a \textbf{Knowledge Distiller} $H$ learns to predict the teacher's multimodal fused representation $\mathbf{z}_{fused}$ from only the mean-pooled image features $\mathbf{z}_{img}$. Thus,
\begin{equation}
    \hat{\mathbf{z}}_{fused} = H(\mathbf{z}_{img}) \in \mathbb{R}^{d_f},
\end{equation}
where $\hat{\mathbf{z}}_{fused}$ is an approximated representation of $\mathbf{z}_{fused}$ (obtained from the teacher). The architecture of the proposed {Knowledge Distiller} is demonstrated in Fig. \ref{fig:kd_architecture}. It is then passed to the frozen shared classifier $\phi_{\text{cls}}$ for WSI-only GA subtype classification.

\subsection{Two-Stage Training}
Our proposed framework employs a two-stage training pipeline.
\justify
\textbf{Stage 1: Multimodal Training.}
 On the teacher side, we train the LMF-based multimodal model using both the image and text modalities by optimizing the cross-entropy loss $\mathcal{L}_{CE}$ between $y$ and $\hat{y}_{T}$. The teacher learns to produce discriminative fused representations $\mathbf{z}_{fused}$ that combine pathological information from WSI patches and diagnostic information from text captions. 
After training, the teacher's module parameters (encoders, LMF module, and classifier) are frozen, as illustrated in Fig. \ref{fig:architecture}. 

\justify
\textbf{Stage 2: Knowledge Distillation for Image-Only Inference.}
Using the teacher's frozen modules, we train the Knowledge Distiller network $H$ to predict the teacher's fused representations directly from image features. The teacher's classifier is shared with the student and kept frozen during this stage. We employ a combined distillation loss, $\mathcal{L}_{distill}$ as follows:
\begin{equation}\label{distill_loss}
    \mathcal{L}_{distill} =
    \mathcal{L}_{CE}
    + \lambda_{feat} 
    \underbrace{\mathcal{L}_{feat}(\hat{\mathbf{z}}_{fused},\, \mathbf{z}_{fused})}_{\text{feature matching loss}}
    + \lambda_{KD}
    \underbrace{\mathcal{L}_{KD}(\hat{\mathbf{y}}_S,\, \hat{\mathbf{y}}_T)}_{\text{logit distillation loss}}
\end{equation}
The \textbf{cross-entropy loss} $\mathcal{L}_{CE}$ supervises the student's predictions $\hat{\mathbf{y}}_S$ against the ground-truth labels $y$. While the \textbf{feature matching loss} uses Smooth L1~\cite{fastrcnn} to maximize the teacher-student agreement in synthesizing the multimodal fused representation by minimizing the distance between the representations predicted by them.
On the other hand, the \textbf{logit distillation loss} uses KL divergence ~\cite{hinton2015} with temperature $\tau$ to align the student's output distribution with the teacher's:
\begin{equation}
    \mathcal{L}_{KD} = \tau^2 \cdot D_{\text{KL}}\!\left(
    \text{softmax}\!\left(\frac{\hat{\mathbf{y}}_S}{\tau}\right)
    \;\Big\|\;
    \text{softmax}\!\left(\frac{\hat{\mathbf{y}}_T}{\tau}\right)
    \right),
\end{equation}
where we experimentally set $\tau = 2$, $\lambda_{feat} = 1.0$, and $\lambda_{KD} = 0.5$.

\justify
\textbf{Why triple supervision?} Cross-entropy alone trains the student's knowledge distillation module from scratch without leveraging the teacher's learned structure. Feature-level supervision enforces the student to reproduce the teacher's intermediate representations, but might not preserve class boundaries if the representation space contains redundant information. Logit-level supervision transfers the teacher's soft class relationships but does not capture the full structure of the fused features. Combining all three ensures the Knowledge Distiller learns representations that are structurally faithful to the teacher's, discriminative for classification, and grounded in the true labels.

\subsection{Inference}
At inference time, the proposed pipeline requires only the WSI, as shown in the student block of the Fig. \ref{fig:architecture}. Using the frozen Phikon~\cite{phikon} encoder $\phi_{img}$, we extract patch-level features, as discussed in Section \ref{image_feature}. We aggregate the patches using mean pooling to obtain a fixed-dimensional WSI-level representation $\mathbf{z}_{img}$, as explained in Section \ref{aggregator}. We pass the $\mathbf{z}_{img}$ through the knowledge distiller $H$ to obtain $\hat{\mathbf{z}}_{fused}$, a sythesized representation of the image-text fused embedding $\mathbf{z}_{fused}$ generated by the teacher. Subsequently, this synthesized representation is then passed through the frozen shared classifier $\phi_{cls}$ to obtain the final predictions, $\hat{\mathbf{y}}_S$, as described in Section \ref{KD}.

Note that the pathology report is used only during training. At inference time, the model requires only the WSI, making it suitable for real-time clinical decision support when reports are not yet available.

\section{Experiments and Results}
\subsection{Dataset}
In our experiments evaluating the proposed method, we use the PatchGastric~\cite{patchgastric} dataset, which originally consists of 991 WSIs, from which patches are extracted and paired with captions from diagnostic reports. There are 262,777 patches in the dataset, each 300x300 pixels in size, extracted at 20x magnification. The dataset contains 9 subtypes of gastric adenocarcinoma, out of which we use 3 following the baselines~\cite{cite,pathm3} for a fair comparison: "well differentiated tubular adenocarcinoma", "moderately differentiated tubular adenocarcinoma" and "poorly differentiated adenocarcinoma". This reduces the patches to 179,285 across 693 WSIs. The data split ratio is the same as in all previous methods. We randomly split the dataset into train (20\%), validate (40\%) and test (40\%). 

\subsection{Implementation Details}
For image feature extraction, we use Phikon~\cite{phikon}, a ViT-B/16 encoder with an input image size of 224×224, in which each input image is split into non-overlapping 16×16 patches. Each WSI patch is resized to 224x224 for input, and then the encoder converts it into a $d_v=768$ dimensional embedding. The final output after passing all the patches from the WSI is $N$ x 768, where $N$ is the number of patches varying across WSIs. For caption encoding we use BGE-small-en-v1.5~\cite{bge}, a retrieval-optimized sentence encoder from the BGE model family. The text encoder returns a $d_t = 384$ dimensional embedding. The LMF module is implemented with the rank $r=128$ (experimentally set as discussed in Section \ref{ra}) to obtain a fused embedding of size $d_f=128$. The shared classifier is a fully connected layer mapping from an input size of 128 to 3, consisting of a ReLU activation and a dropout layer with probability 0.3. The teacher model is trained for 20 epochs using the Adam~\cite{adam} optimzer with a learning rate of 1e-4, optimizing cross-entropy loss. 

\textcolor{black}{As illustrated in Fig. \ref{fig:kd_architecture}, the Knowledge Distiller (KD) is implemented as a residual network [25] consisting of three components. First, an input projection maps a $d_v=768$ dimensional input to $d_h = 512$ via a linear layer followed by Layer Normalization [26], GELU [27] activation, and dropout with probability 0.3. Second, each residual block preserves the $512$-dimensional hidden representation and consists of two ($512 \rightarrow 512$) linear layers, each followed by Layer Normalization and GELU activation, with a skip connection between the block input and output. A dropout layer with a probability of 0.4 is applied within each residual block. Third, the hidden representation $d_h = 512$ is then mapped to the output $d_f = 128$ via a linear layer followed by Layer Normalization [26].} The student model is trained for 50 epochs using the Adam~\cite{adam} optimizer with a learning rate of 1e-4, optimizing the distillation loss $\mathcal{L}_{distill}$ defined in \eqref{distill_loss}.

The proposed method is implemented in PyTorch, and all experiments are evaluated on a computing system with NVIDIA RTX 6000 ADA 48GB GPU, 512 GB of RAM and an Intel Xeon Gold 6426Y processor. Reported classification accuracy is the mean and standard deviation over three independent runs.
\begin{table}[t]
\caption{Comparison with state-of-the-art (SOTA) methods under image-only inference. Results are reported as mean accuracy (\%) $\pm$ standard deviation. $\boldsymbol{\Delta}$ denotes the difference in mean accuracy of SOTA methods compared to our method (MKD-LMF). \textcolor{red}{Red-colored} negative values indicate lower performance than ours, while positive values indicate higher performance.}
\label{tab:comparison}
\centering
\begin{tabular}{l|c|c}
\hline
\textbf{Method} & \textbf{Accuracy (\%)} & $\boldsymbol{\Delta}$ \\
\hline
ABMIL~\cite{abmil} & $66.67 \pm 2.63$ & \color{red}$-8.16$ \\
CLAM~\cite{clam} & $67.83 \pm 1.12$ & \color{red}$-7.00$ \\
DSMIL~\cite{dsmil} & $69.02 \pm 0.42$ & \color{red}$-5.81$ \\
TransMIL~\cite{transmil} & $67.48 \pm 1.16$ & \color{red}$-7.35$ \\
CITE~\cite{cite} & $69.63 \pm 0.91$ & \color{red}$-5.20$ \\
ILRA-MIL~\cite{ilramil} & $70.16 \pm 1.11$ & \color{red}$-4.67$ \\
PathM3~\cite{pathm3} & $71.48 \pm 1.30$ & \color{red}$-3.35$ \\ \hline
\color{blue}\textbf{MKD-LMF (Ours)} & \color{blue}$\mathbf{74.83 \pm 1.49}$ &  \\
\hline
\end{tabular}
\end{table}

\subsection{Results and Analysis}\label{ra}
\justify
\textbf{Comparison with state-of-the-art methods.}
We compare our method against state-of-the-art WSI classification methods under the image-only inference setting. Results for all baseline methods are taken from PathM3~\cite{pathm3}, which uses identical experimental settings, including the same dataset, data split ratio, and three GA subtypes for classification, thereby enabling direct comparison.
As shown in Table~\ref{tab:comparison}, our method achieves 74.83\% accuracy, outperforming all baselines, including PathM3 (71.48\%) by {3.35} percentage points and the best attention-based MIL method ILRA-MIL (70.16\%) by {4.67} percentage points. The consistent margin over established MIL baselines, including ABMIL, CLAM, DSMIL, TransMIL, and CITE, demonstrates the effectiveness of our approach in gastric adenocarcinoma subtype classification.

\justify
\textbf{Ablation study.}
\begin{table}[t]
\caption{Ablation study of the proposed MKD-LMF framework. The results of the experiments in the table are reported for a single run. The table analyzes (i) the performance of the teacher model using both image and text during inference, which is not clinically relevant, (ii) the effect of different distillation loss components in combination with the classification loss (cross-entropy) on the student-side and (iii) the impact of different inference modalities. Results show that combining both losses yields the best student performance. The student model in our multimodal knowledge distillation (KD) framework substantially outperforms unimodal (image-only or text-only) inference settings.}
\label{tab:ablation}
\centering
\begin{tabular}{p{3.2cm}|p{6.4cm}|c}
\hline
\textbf{Study} & \textbf{Configuration} & \textbf{Accuracy (\%)}\\
\hline
Multimodal Inference &
Proposed teacher model &
$\mathbf{92.74} $\\
\hline
\multirow{3}{*}{Distillation Loss}
& Feature loss ($\mathcal{L}_{\text{feat}}$) & $73.63$\\
& KD loss ($\mathcal{L}_{\text{KD}}$) & $73.13$\\
& Feature loss + KD loss & $\mathbf{74.83}$\\
\hline
\multirow{3}{*}{Inference Modality}
& Text-only (unimodal) & $68.76$\\
& Image-only (unimodal) & $70.35$\\
& Image-only (proposed multimodal KD)  & $\mathbf{74.83}$\\
\hline
\end{tabular}
\end{table}
We conduct ablation studies to validate the key design choices in our framework, as described in the following three paragraphs. Notably, results of our ablation study are summarized in Table \ref{tab:ablation}, where accuracy is determined in only one run of each model. 

\paragraph{Multimodal Inference.} We evaluated the teacher model independently under a multimodal inference setting using both image and text inputs. The teacher achieves the best performance among all ablated variants (see Table \ref{tab:ablation}) by leveraging complementary information from the whole-slide image and the diagnostic report summary to classify GA subtypes. However, as discussed in Section 1, this multimodal setup is not clinically practical, since report generation requires clinician involvement and is assumed to be not available at inference time.

\paragraph{Importance of Distillation Loss Components.}
We ablate the proposed dual-objective distillation loss by training the student model with each component independently and in combination, while keeping the cross-entropy loss fixed. Using only the feature loss $(\mathcal{L}_{\text{feat}})$ or only the KD logit loss $(\mathcal{L}_{\text{KD}})$ leads to lower performance, lagging behind the combined objective by $1.20\%$ and $1.70\%$ points, respectively. In contrast, jointly optimizing both losses yields the best performance, demonstrating that they provide complementary supervision signals for more effective knowledge transfer.

\paragraph{Unimodal vs. Image-Only Multimodal KD.}
To quantify the contribution of multimodal fusion, we evaluate classifiers trained on individual modalities without fusion or KD. The image-only unimodal model achieves an accuracy of 70.35\%, whereas the text-only model obtains 68.76\%. Compared with these unimodal baselines, the distilled student model (i.e., our image-only multimodal KD framework) improves the performance by {4.48 percentage points} over the image-only model and by {6.07 percentage points} over the text-only model. Furthermore, the full multimodal teacher model achieves a substantial gain of {17.91 percentage points} over the distilled student model. These results demonstrate that the LMF-based fusion effectively leverages complementary information across multiple modalities and that the Feature Distiller successfully transfers this multimodal knowledge to our image-only inference pipeline.

\begin{figure}[t]
    \centering
    \begin{subfigure}[b]{0.45\linewidth}
        \centering
        \includegraphics[width=\linewidth]{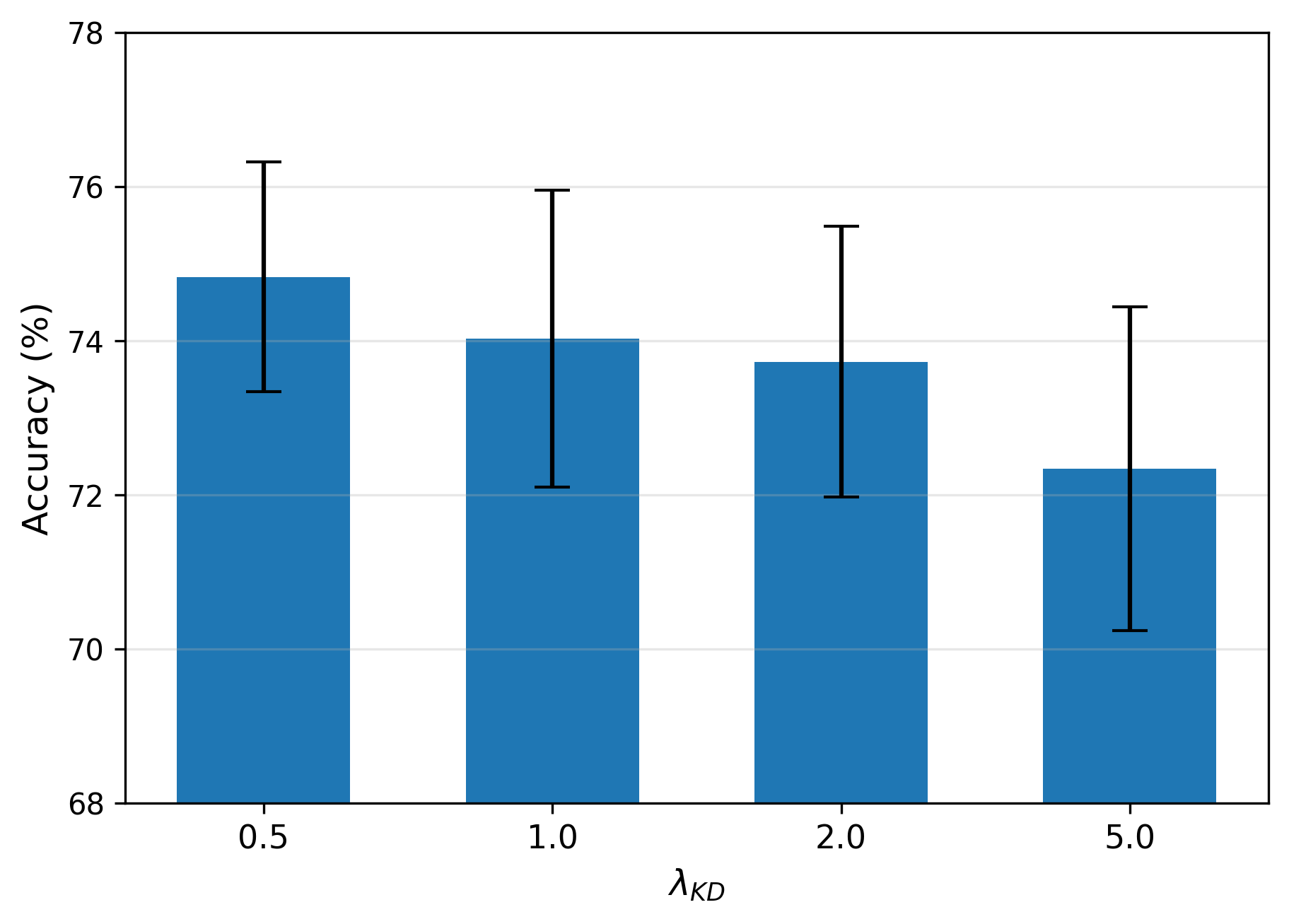}
        \caption{Effect of $\lambda_{KD}$.}
        \label{fig:lamb_kd}
    \end{subfigure}
    \begin{subfigure}[b]{0.45\linewidth}
        \centering
        \includegraphics[width=\linewidth]{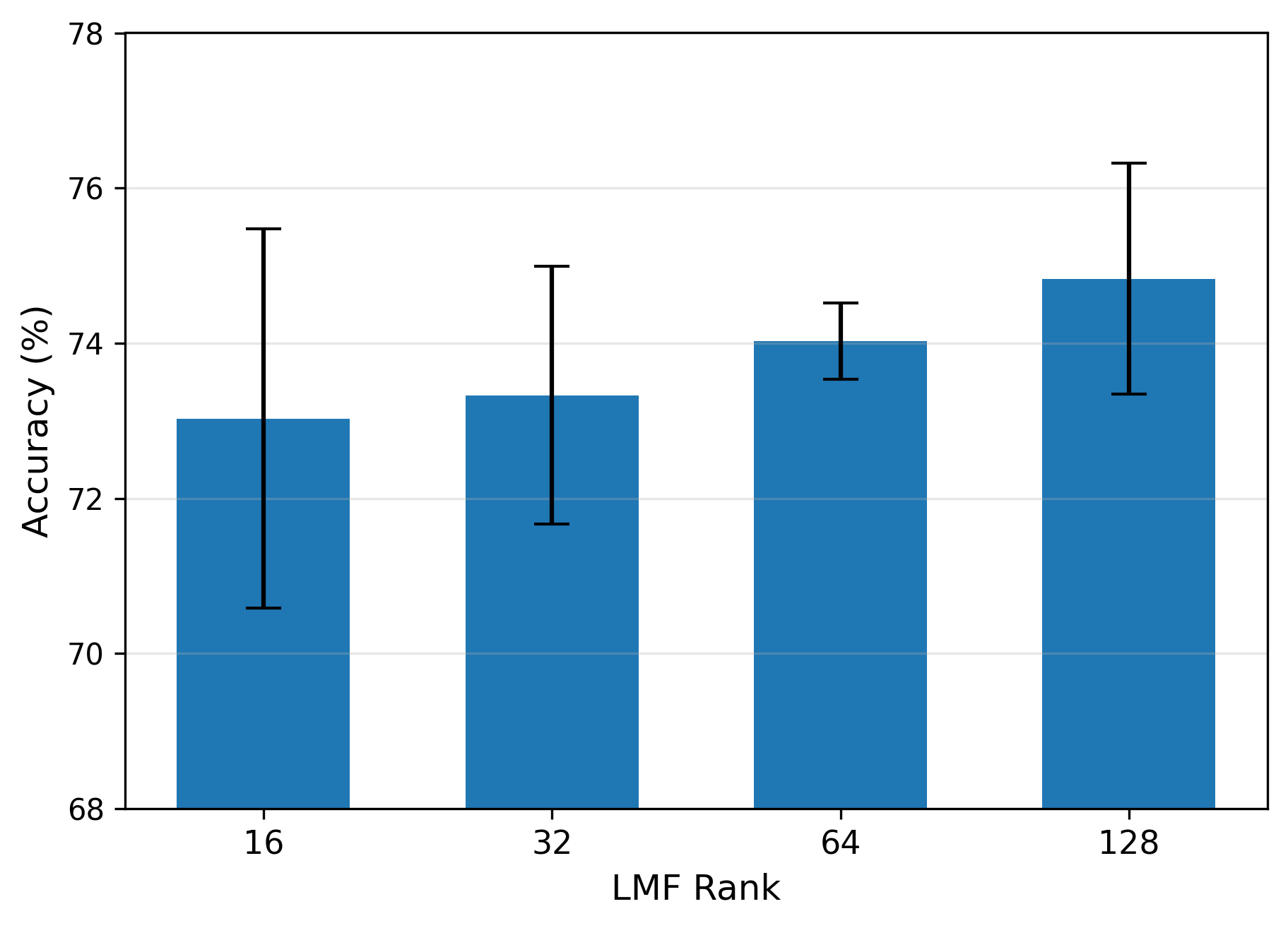}
        \caption{Effect of LMF rank.}
        \label{fig:rank}
    \end{subfigure}
    \caption{Results for different choices of (a) the coefficient of logit distillation loss, $\lambda_{KD}$ and (b) the rank in the low-rank multimodal fusion (LMF) approach}
    \label{fig:ablation}
\end{figure}

\justify
\textbf{Choice of logit distillation loss coefficient.}
We also experimented with tuning the $\lambda_{KD}$ parameter to four settings (0.5, 1.0, 2.0, \& 5.0). Our best results (74.83\%) were achieved with $\lambda_{KD}$ = 0.5. The accuracy of training the student with other values is presented in Figure \ref{fig:lamb_kd}.

\justify
\textbf{Choice of rank in LMF.}
We experimented with the teacher model's rank hyperparameter for implementing LMF. The student model distils knowledge from the teacher model, which is trained with four different ranks: 16, 32, 64, and 128. The highest accuracy was observed with $r = 128$, as shown in Fig. \ref{fig:rank}. It is likely that a higher rank enables the teacher to learn richer multimodal representations via cross-modal interactions.
\begin{figure}[t]
\centering
\begin{subfigure}[t]{0.24\textwidth}
    \centering
    \includegraphics[width=\linewidth]{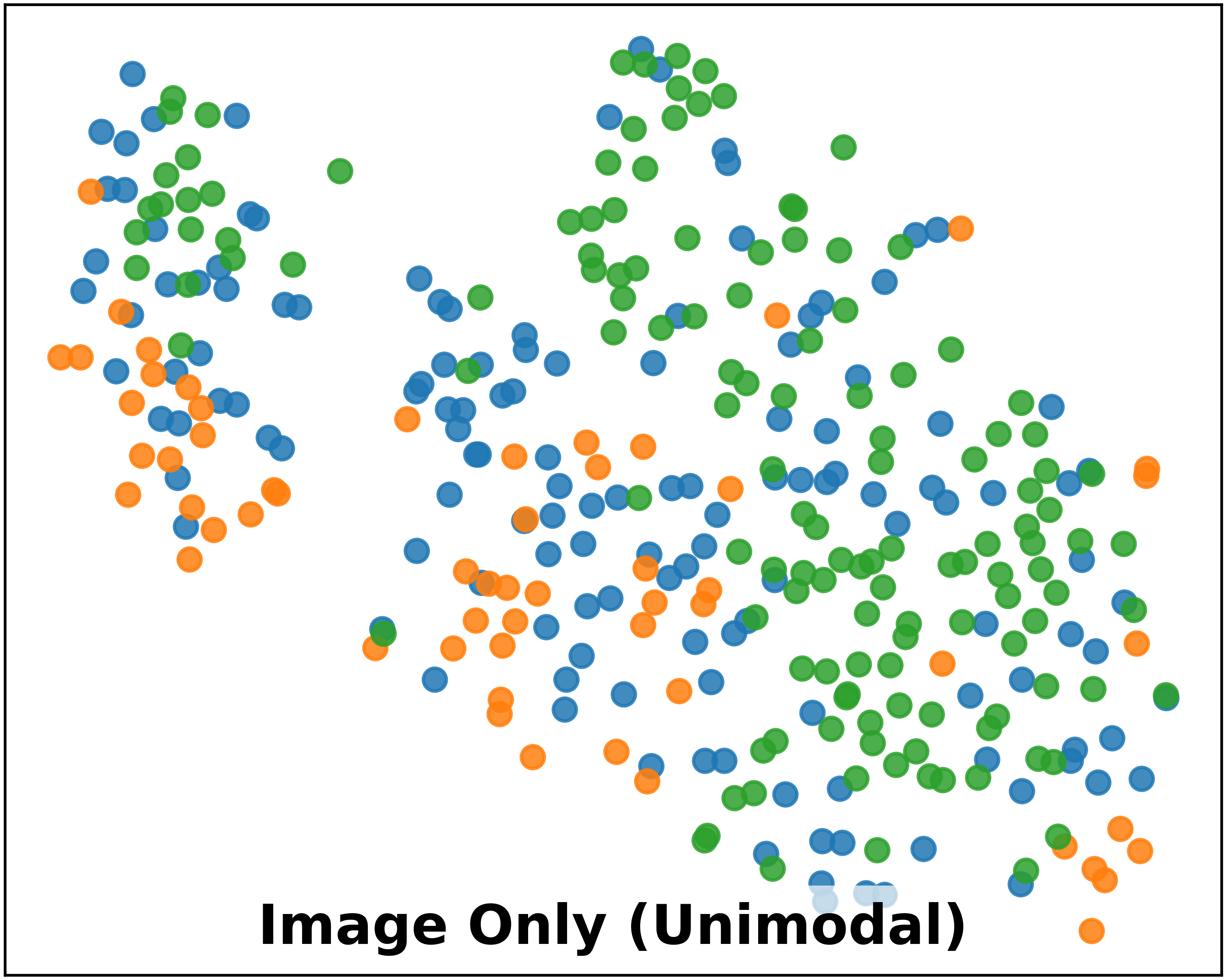}
    \label{fig:tsne_img}
\end{subfigure}
\begin{subfigure}[t]{0.24\textwidth}
    \centering
    \includegraphics[width=\linewidth]{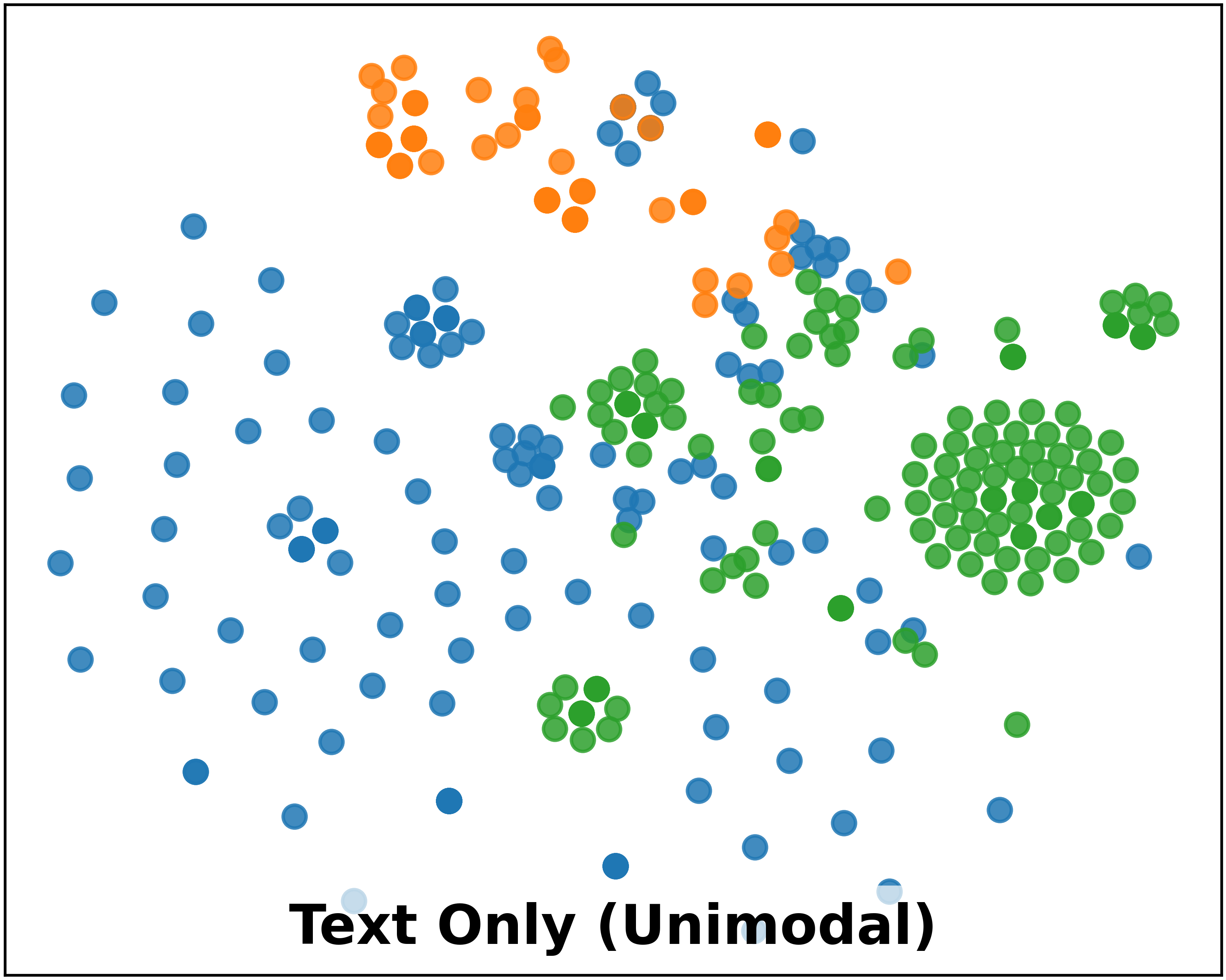}
    \label{fig:tsne_text}
\end{subfigure}
\begin{subfigure}[t]{0.24\textwidth}
    \centering
    \includegraphics[width=\linewidth]{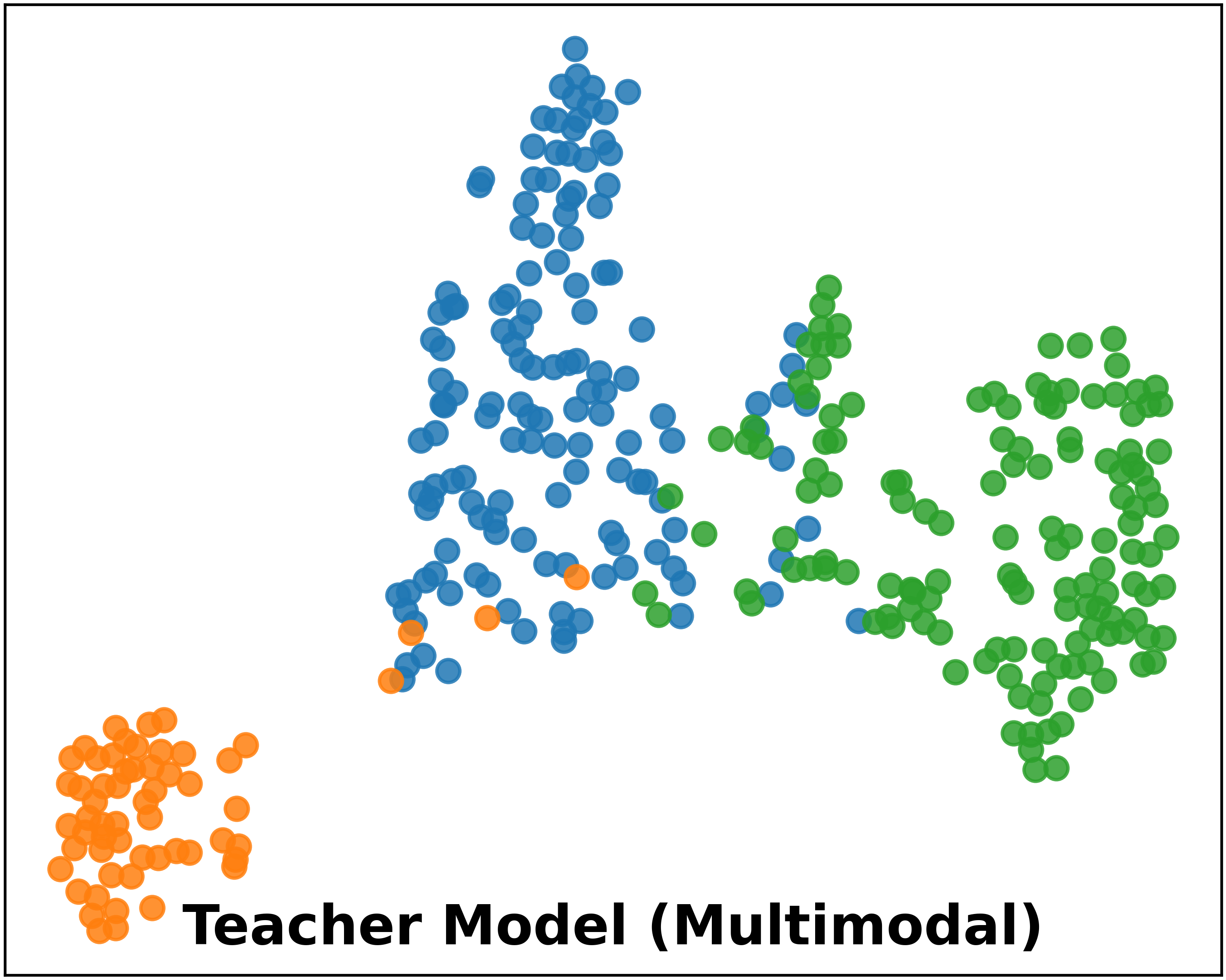}
    \label{fig:tsne_teacher}
\end{subfigure}
\begin{subfigure}[t]{0.24\textwidth}
    \centering
    \includegraphics[width=\linewidth]{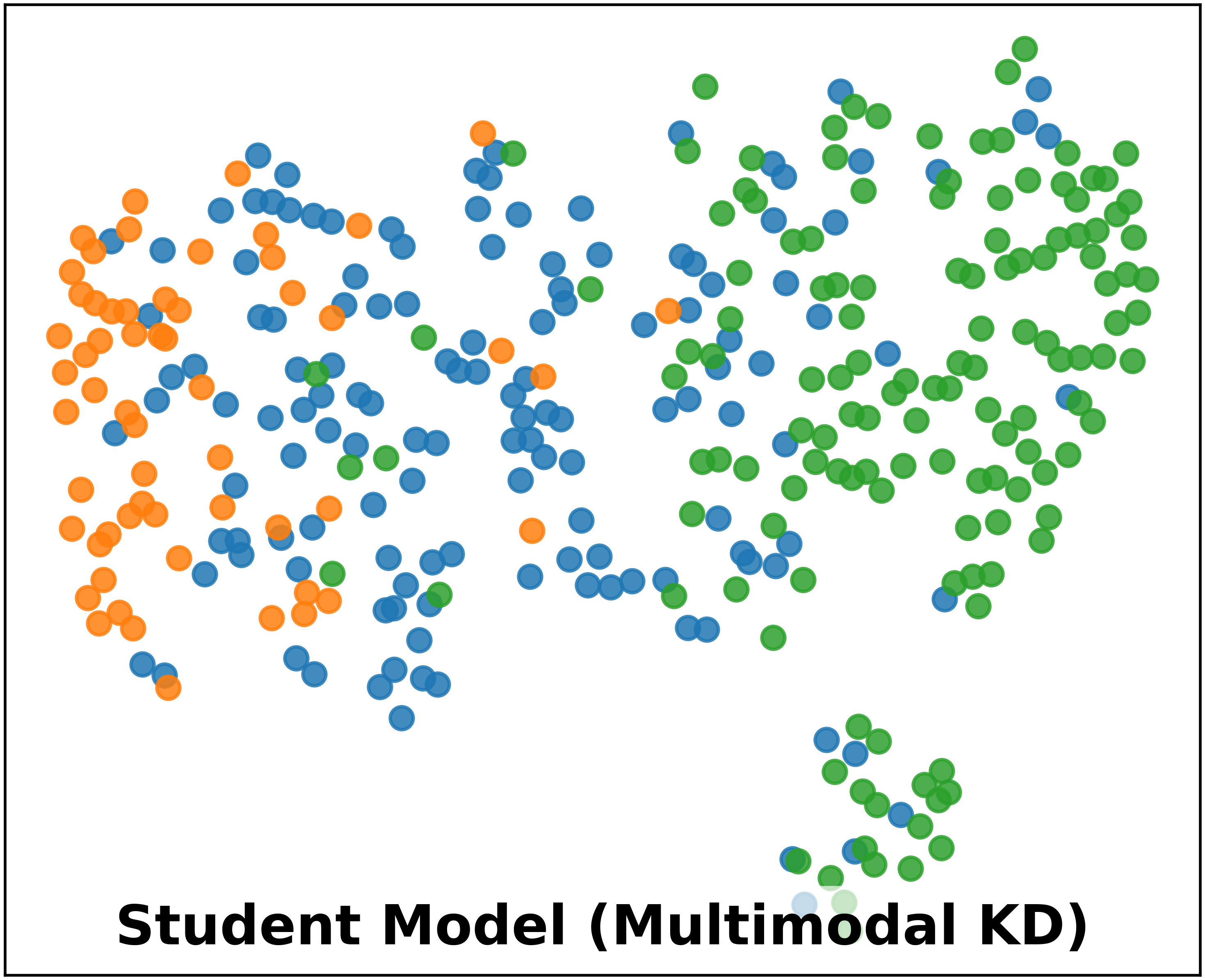}
    \label{fig:tsne_student}
\end{subfigure}
\caption{
t-SNE visualization of learned embeddings with corresponding DBI scores: image-only (5.2641), text-only (2.3107), teacher (0.7309), and distilled student (1.1078). \textcolor{blue}{Blue} represents "moderately differentiated tubular adenocarcinoma". \textcolor{orange}{Orange} represents "poorly diffrentiated adenocarcinoma". \textcolor{green}{Green} represents "well diffrentiated tubular adenocarcinoma"
}
\label{fig:tsne}
\end{figure}

\justify
\textbf{Feature visualizations.} Fig.~\ref{fig:tsne} shows t-SNE plots of learned representations which are $\mathbf{z}_{img}$ (image-only features), $\mathbf{z}_{txt}$ (text-only features), $\mathbf{z}_{fused}$ (LMF representation of fusion of text and image embeddings) and $\hat{\mathbf{z}}_{fused}$ (Predicted embeddings by the knowledge distiller from the patch features only). To qualitatively assess the representation quality, we compute the Davies-Bouldin Index (DBI) on the learned embeddings. It measures the ratio of within-cluster dispersion to between-cluster separation, where lower values indicate better-defined clusters. The teacher model achieves the lowest DBI (0.7309), followed by the distilled student model (1.1078), both substantially outperforming the text-only (2.3107) and image-only (5.2641) representations, confirming the improved cluster compactness and inter-class separation.

\section{Conclusion}
We proposed an LMF-based multimodal fusion framework for image-only classification of gastric adenocarcinoma, fusing diagnostic captions with patches extracted from histopathological whole-slide images (WSIs). The proposed method outperforms the strongest baseline by 3.35\%. Our results suggest that carefully chosen fusion mechanisms can outperform architecturally complex methods in computational pathology. While our framework demonstrates strong performance, it is limited to a single dataset and a single cancer type. Future work could explore evaluation across diverse datasets of various cancer types.

\end{document}